\documentclass{article}
\usepackage{spconf,amsmath,graphicx,hyperref}
\usepackage{enumitem}
\usepackage{multirow}
\usepackage{floatrow}
\usepackage{xcolor}
\usepackage{colortbl}
\usepackage{subfigure}

\newcommand{\repourl}{\url{https://github.com/hcai-mms/MSM-face-voice}}

\definecolor{Gray}{gray}{0.95}
\definecolor{Light}{gray}{0.98}
\definecolor{lblue}{rgb}{0.94,0.96,0.99}

\newcommand{\facevector}{\mathbf{f}}
\newcommand{\voicevector}{\mathbf{v}}
\newcommand{\multimodalvector}{\mathbf{m}}

\newfloatcommand{capbtabbox}{table}[][\FBwidth]

\title{Face-Voice Association with Inductive Bias for \\Maximum Class Separation}
\name{
\begin{tabular}{c}
Marta Moscati$^{1}$, Oleksandr Kats$^{1}$, Mubashir Noman$^{2}$, Muhammad Zaigham Zaheer$^{2}$ \\  Yufang Hou$^{3}$, Markus Schedl$^{1,4}$, Shah Nawaz$^{1}$
\end{tabular}}
\address{
$^{1}$Johannes Kepler University Linz, Austria,
$^{2}$MBZUAI, UAE \\
$^{3}$IT:U Interdisciplinary Transformation University Austria, 
$^{4}$Linz Institute of Technology, Austria
}
\begin{document}
%
\maketitle
\begin{abstract}
    Face-voice association is widely studied in multimodal learning and is approached representing faces and voices with embeddings that  are close for a same person and well separated from those of others. Previous work achieved this with loss functions. Recent advancements in classification have shown that the discriminative ability of embeddings can be strengthened by imposing maximum class separation as inductive bias. This technique has never been used in the domain of face-voice association, and this work aims at filling this gap. More specifically, we develop a method for face-voice association that imposes maximum class separation among multimodal representations of different speakers as an inductive bias. Through quantitative experiments we demonstrate the effectiveness of our approach, showing that it achieves SOTA performance on two task formulation of face-voice association. Furthermore, we carry out an ablation study to show that imposing inductive bias is most effective when combined with losses for inter-class orthogonality. To the best of our knowledge, this work is the first that applies and demonstrates the effectiveness of maximum class separation as an inductive bias in multimodal learning; it hence paves the way to establish a new paradigm\footnote[1]{Code: \repourl}.
\end{abstract}

\section{Introduction}
    \label{sec:intro}
       \vspace{-0.7em}
    Face-voice association aims at associating faces and voices of same persons~\cite{kamachi2003putting}. When devised as a machine learning task, face-voice association is formulated as a multimodal learning task, since faces are captured visually and voices are captured by audio signals. From its formulation as \emph{cross-modal verification}, i.e. determining whether a face and a voice belong to the same person, and \emph{cross-modal matching}~\cite{nagrani2018seeing}, i.e. identifying the face out of a set of candidates that belongs to the same person of a given voice, several methods have been proposed to address this task.
    These methods rely on representing faces and voices in a shared embedding space, where representations of faces and voices are close if they belong to the same person
    ~\cite{nagrani2018learnable,nawaz2019deep, saeed2022fusion,wen2018disjoint}.
    Some approaches~\cite{nagrani2018seeing,nagrani2018learnable,horiguchi2018face} model the distances explicitly in the loss function - e.g. by means of contrastive or triplet losses that aim at reducing distances between matching pairs. Although effective, these approaches do not scale well to cases where the number $n$ of faces and voices  is large; this is due to their training complexities of orders $O(n^2)$ or $O(n^3)$ and results in long training time as the size of the dataset increases.
    Other approaches~\cite{nawaz2019deep, saeed2022fusion} propose the use of losses for classification 
    to improve the discriminative power of the learned embeddings by minimizing intra-class and maximizing inter-class distances, defining classes in terms of speakers. However, these losses alone lack the inherent discriminative capability for learning separable embeddings~\cite{saeed2022fusion}. To mitigate this, several prior works proposed enforcing orthogonality constraints along with classification losses to improve the discriminative power of the learned embeddings by introducing additional loss terms~\cite{li2020oslnet,ranasinghe2021orthogonal,saeed2022fusion,hannan2025paeff}. 
    
    Recently, several works on classification tasks~\cite{yang2022neural_collapse,kasarla2022maximum} showed that maximum class separation can be incorporated as an inductive bias by adding a fixed matrix multiplication before computing the final softmax activations, with the advantage of an improved classification performance.
    Though effective, so far these methods have only been applied to unimodal classification tasks; in other words, they have never been applied to multimodal learning scenarios \textit{nor} to tasks beyond classification. 
    To fill this gap in the current status of research, we propose a framework for face-voice association that incorporates maximum class separation as an inductive bias by adding a fixed matrix to the network prior to training. The network parameters -- excluding the inductive-bias matrix -- are optimized by means of orthogonality constraints that minimize the angular distance between faces and voices of same speakers and maximize the distance between different speakers. We demonstrate that enforcing maximum class separation among faces and voices of speakers and by means of a fixed inductive-bias matrix provides notable improvements on face-voice association, both as cross-modal verification and matching. In summary, our contributions are the following:
    
\begin{figure*}[t]
    \centering
    \includegraphics[width=0.8\linewidth]{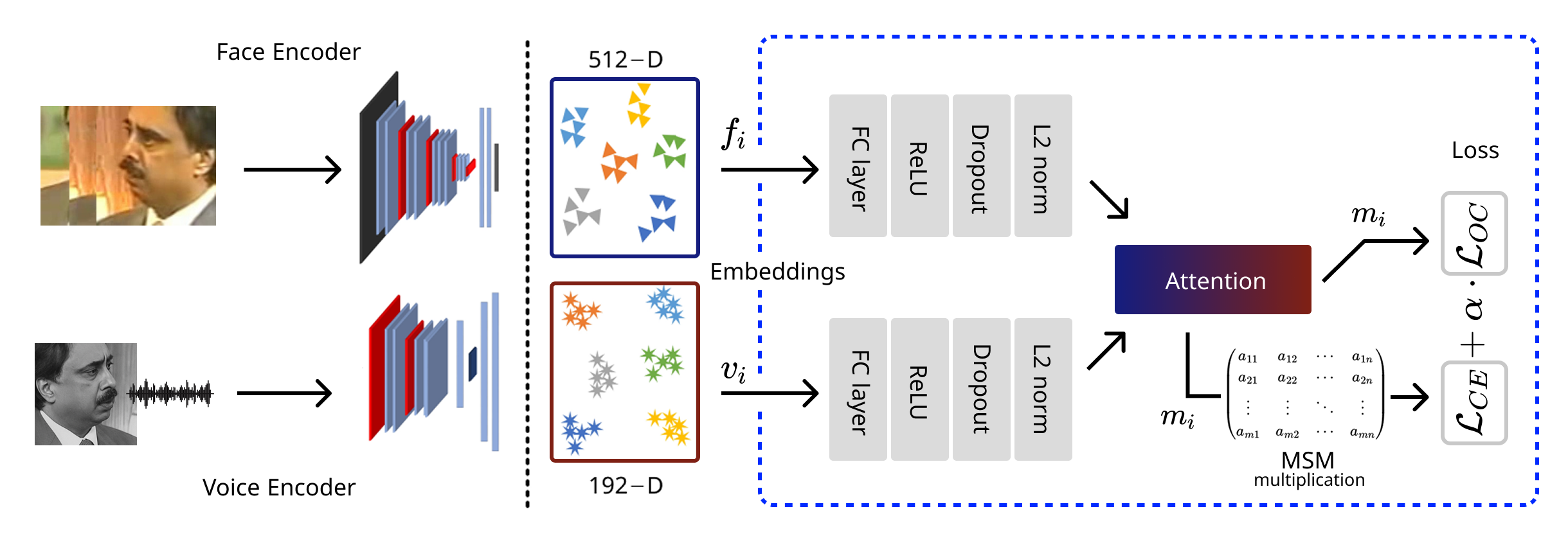}
    \vspace{-1.7em}
    \caption{Overall illustration of the proposed face-voice association method.  
    }
    \label{fig:network}
\end{figure*}

    \begin{itemize}[leftmargin=2mm,topsep=0pt,itemsep=-1pt]
        \item We propose an approach 
        that uses a fixed inductive-bias matrix to enforce maximum class separation among multimodal representations of different speakers.
        \item We demonstrate the effectiveness of our proposed approach by means of quantitative experiments on two benchmark datasets for face-voice association, showing that our approach outperforms existing methods on both tasks of cross-modal verification and cross-modal matching.   
        \item By means of an ablation study, we show 
        that imposing the inductive bias is most effective when combined with losses for inter-class orthogonality.
    \end{itemize}

\vspace{-2mm}
\section{Method}
\label{sec:method}
   \vspace{-0.7em}
    \noindent \textbf{Preliminaries.}
        We denote with $i\in \mathcal{V}$ the $i$-th \textit{instance}, e.g. a video snippet of one speaker, and with $s_i\in \mathcal{S}$ the speaker of $i$. Since speakers appear in more than one instance, the number of speakers $N_s = |\mathcal{S}|$ is smaller than 
        the number of instances $N_s < |\mathcal{V}|$. We denote with $\facevector_i$, $\voicevector_i$, and $\multimodalvector_i$ respectively to the representations of the face, voice, and multimodal representation (i.e. face \textit{and} voice) from $i$. Given a face-voice pair $(\facevector_i, \voicevector_j)$, the task of face-voice association as \textit{cross-modal verification} consists in determining whether face $\facevector_i$ and voice $\voicevector_j$ belong to the same speaker, i.e. whether $s_i=s_j$. %
        In contrast, \textit{cross-modal matching} involves comparing a probe (input modality) against a gallery of elements of 
        the other modality
        . More concretely, given a voice $\voicevector_i$ as probe and a gallery $\{\facevector_j | \exists! j : s_j=s_i \}_{j=1}^{n_c}$ of $n_c$ faces such that only one corresponds to the same speaker of the voice, cross-modal matching consists in identifying the unique face $\facevector_j$ in the gallery such that $s_i=s_j$. The task of cross-modal matching can be similarly formulated from face to voice matching. 

    \noindent \textbf{Model Architecture.}
        Figure~\ref{fig:network} provides an overview of the model architecture. Face and voice are processed by two distinct branches: given an instance $i$, face is extracted encoding one visual frame of the video with a pre-trained FaceNet~\cite{schroff2015facenet} (Face Encoder in Figure~\ref{fig:network}), while voice is extracted encoding the audio signal with a pre-trained Ecapa-tdnn~\cite{desplanques2020ecapa} (Voice Encoder in Figure~\ref{fig:network}). These two feature extraction modules are kept frozen and their parameters are not updated during model training. The resulting embeddings are passed through two separate networks, each consisting of a fully connected layer of output dimension $d=128$ with ReLU as non-linear activation function, followed by a dropout layer and $L^2$ normalization. The resulting vectors are $\facevector_i, \voicevector_i\in \mathrm{R}^d$. We combine $\facevector_i$ and $\voicevector_i$ into a multimodal representation $\multimodalvector_i$ by taking the weighted average with weight adjusted during training. 

        To apply maximum class separation as an inductive bias in a way that follows the formulation proposed by Kasarla et al.~\cite{kasarla2022maximum},
        , we set the dimension of $\multimodalvector_i$ to $N_s - 1$, where $N_s$ represents the number of speakers. We then enforce maximum class separation as inductive bias by multiplying $\multimodalvector_i$ with the maximally-separated matrix $P_{N_s-1}\in \mathrm{R}^{(N_s-1)\times N_s}$, constructed recursively starting from the vector $P_1 = (1, -1)$ and as:
        \begin{equation}
            P_{N_s-1} = \begin{pmatrix}
                            1 & -\frac{1}{N_s-1} \mathbf{1}\\
                            \mathbf{0} & \sqrt{1-\frac{1}{(N_s-1)^2}}P_{N_s-2}
                        \end{pmatrix}\in\mathrm{R}^{(N_s-1)\times N_s},
        \end{equation}  
        where $\mathbf{0}$ and $\mathbf{1}$ denote vectors consisting of $0$'s and $1$'s, respectively. The vector of logits over speakers will therefore be given by $\hat{\mathbf{c}}_i = P_{N_s-1} \cdot \multimodalvector_i \in \mathrm{R}^{N_s}$.

    \noindent \textbf{Training.}
        During training, batches $B=\{((\facevector_i, \voicevector_i), s_i)\}$ consist of pairs of faces and voices $(\facevector_i, \voicevector_i)$ of a same instance $i$, and of the corresponding speaker label $s_i$. 
        Similar to Saeed et al.~\cite{saeed2022fusion}, we utilize two loss functions: one loss that pushes the model to correctly identify the speaker $s_i$ of each instance $i$ within the batch, and one loss that pushes representations of entities of the same speaker to be aligned, and of different speakers to be orthogonal to each other. We use cross-entropy $\mathcal{L}_\text{CE}$ with $N_s$ classes as speaker classification loss; for a single instance $i$, this is given by:
        \begin{equation}
            \label{eq:cross_entropy}
            \mathcal{L}_{\text{CE}, i} = - \log \frac{\exp(\hat{{c}}_{i,s_i})}{\sum_{j=1}^{N_s}\exp(\hat{c}_{i,j})},
        \end{equation}
        where $\hat{{c}}_{i,j}$ represents the component $j$ of the vector of logits  $\hat{\mathbf{c}}_i$, i.e. the score assigned by the model for instance $i$ to speaker $j$. The index $s_i$ represents the component of the correct speaker of instance $i$; therefore $\hat{{c}}_{i,s_i}$ represents the score assigned by the model to the correct speaker. The cross-entropy loss for the batch is the sum over all instances $i$ within the batch. We furthermore use the orthogonal constraints loss $\mathcal{L}_\text{OC}$, defined as~\cite{saeed2022fusion}:
        \begin{equation}
            \label{eq:orthogonal_constraints_loss}
            \mathcal{L}_\text{OC} = 1 - \sum_{i,j\in B, s_i = s_j} \cos({\multimodalvector_i, \multimodalvector_j}) + \left|\sum_{i,k\in B, s_i \neq s_k} \cos({\multimodalvector_i, \multimodalvector_k})\right|,
        \end{equation}
        where $\cos$ represents the cosine similarity; the first sum runs over pairs of instances $i, j$ corresponding to same speakers, while the second to pairs corresponding to different speakers.
        
        The overall training objective is given by $\mathcal{L} = \mathcal{L}_\text{CE} +\alpha \mathcal{L}_\text{OC}$, where $\alpha$ is a hyperparameter chosen to optimize the model performance on the validation set.

   
    \begin{table}
        \centering
        \resizebox{0.95\linewidth}{!}{
        \setlength{\tabcolsep}{14pt}
        \begin{tabular}{l|cc|cc}
            \rowcolor{Gray}
            \hline
            Method  &  EER $\downarrow$ & AUC $\uparrow$ & EER $\downarrow$ & AUC $\uparrow$ \\
            \cline{2-5}
            \rowcolor{Gray}
             & \multicolumn{2}{c|}{Seen-Heard} & \multicolumn{2}{c}{Unseen-Unheard} \\
            \hline
            \hline
            DIMNet~\cite{wen2018disjoint}              & --    & --      & \underline{24.9}     & --          \\
            Learnable Pins~\cite{nagrani2018learnable} & 21.4 & 87.0   & 29.6          & 78.5        \\
            Single Stream Net.~\cite{nawaz2019deep}    & \underline{17.2} & 91.1          &  29.5 & 78.8  \\
            AML~\cite{zheng2021adversarial}            & --    & \underline{92.3 }  & --             &  80.6            \\
            DRL~\cite{ning2021disentangled}            & --   & --      & 25.0          &  84.6             \\
            FOP~\cite{saeed2022fusion}                 & 19.3 & 89.3   & 24.9 & 83.5      \\
            SBNet.~\cite{saeed2023single}              & --    & --      & 25.7 & 82.4      \\
            Wang et al.~\cite{wang2020learning}        & --    & --      & --    & \textbf{85.0} \\ 
            VF Aligner~\cite{jiang2024target}          & --    & --      & \underline{24.4}  &  83.0 \\
            \rowcolor{lblue}
            \hline
            Ours                                       & \textbf{13.9} & \textbf{93.7} & \textbf{22.9} & \textbf{85.0}      \\
            \hline
        \end{tabular}
        }
        \caption{Cross-modal verification results on \textit{seen-heard} and \textit{unseen-unheard} configurations 
        of VoxCeleb
        . 
        Combinations of model and configuration for which results are not available are marked with --. 
        Bold: best, underlined: second best. 
        }
        \vspace{-1.0em}
        \label{tab:sota}
    \end{table}

    \noindent \textbf{Inference.}
        Once the model is trained, at inference time we evaluate its performance on two formulations: \textit{cross-modal verification} and \textit{cross-modal matching}.
        
        \noindent \textit{Cross-Modal Verification.} %
            Given a test set pair $\{(\facevector_i, \voicevector_j)\}$ of face and voice, cross-modal verification consists in binary classifying whether they belong to the same speaker, i.e. $s_i=s_j$. Model assigns a score to each pair according to the cosine similarity between face and voice, $\cos(\facevector_i, \voicevector_i)$; the pair is classified as positive ($s_i=s_j$) if the score exceeds a threshold.
            
        \noindent \textit{Cross-Modal Matching.} %
            We formulate cross-modal matching in the case of having voice as input and face as target; the task can be analogously formulated with face as input. The test-set consists of the voice $\voicevector_i$ of one instance $i$, and a gallery of $n_c$ faces $\{\facevector_j\}_{j=1}^{n_c}$. The speaker of one of the $n_c$ faces of the gallery is the same of the input instance, i.e. $\exists! j : s_j=s_i$. Cross-modal matching consists in identifying the unique instance in the gallery that matches the input speaker, $s_i=s_j$. To each pair $\{(\voicevector_i, \facevector_j)\}_{j=1}^{n_c}$ our model assigns a score $\cos(\voicevector_i, \facevector_j)$ given by the cosine similarity of the voice and face representations. The face with the highest similarity is identified as the matching face.
            \vspace{-2mm}

\section{Experimental Setup}
    \label{sec:experiments}
       \vspace{-0.7em}
    \noindent \textbf{Implementation Details.}
    In our experiments, we utilize one Quadro RTX $6000$ GPU. Our model is trained for $50$ epochs using a batch-size of $512$. 
    We use Adam optimizer with exponentially decaying learning rate and set the initial learning rate to $10^{-4}$. 
    We employ FaceNet~\cite{schroff2015facenet} and Ecapa-tdnn~\cite{desplanques2020ecapa} encoders to extract face and voice feature embeddings. 
    
    \noindent \textbf{Datasets.} 
    We perform experiments on VoxCeleb and MAV-Celeb, benchmark datasets for face-voice association. VoxCeleb~\cite{nagrani2017voxceleb} is an audio-visual dataset consisting of more than $100,000$ short clips extracted from over $20,000$ interview videos of $1,251$ speakers. We used the same train, validation, and test splits used in prior works~\cite{nagrani2018learnable} to evaluate the performance of models on seen-heard and unseen-unheard configurations. MAV-Celeb~\cite{nawaz2021cross,saeed2024synopsis} was introduced to study the impact of languages on face-voice association with cross-modal verification task. 
    It consists of videos of celebrities in different languages (e.g. Eng. and Urdu) of $70$ distinct speakers. 
    We use the same train and test splits used in prior works to evaluate the performance on Urdu and English splits~\cite{tao2024multi,tang2024exploring,moscati2025linking}.

\begin{table}[t]
    \centering
    \scalebox{0.80}{
    \setlength{\tabcolsep}{12pt}
        \begin{tabular}{llcc}
        \rowcolor{Gray}
        \hline
        Method & Language & (EER $\downarrow$)  & All (EER $\downarrow$)  \\
        \hline
        
        \multirow{2}{*}{FOP~\cite{saeed2022fusion}} & English      & 29.3  & \multirow{2}{*}{27.5} \\
        \cline{2-3}
        
        & Urdu         & 25.8  &  \\
        \hline
        
        \multirow{2}{*}{Xaiofei~\cite{tang2024exploring}}& English      & 28.5   & \multirow{2}{*}{24.7}   \\
        \cline{2-3}
        
        & Urdu         &  20.9  &    \\
        \hline
        
        \multirow{2}{*}{Audio-visual~\cite{chen2024contrastive}} & English & 17.1  & \multirow{2}{*}{\textbf{17.7}} \\
        \cline{2-3}
         & Urdu         & 18.4  &   \\
        \hline
        
        \multirow{2}{*}{HLT~\cite{tao2024multi}} & English  & 21.8   & \multirow{2}{*}{\underline{18.2}}  \\
        \cline{2-3}

        & Urdu  & 14.7   &     \\
        \hline
        
        \rowcolor{lblue}
        & English & 16.5  &   \\
        \cline{2-3}
        
        \cellcolor{lblue}\multirow{-2}{*}{Ours}  & \cellcolor{lblue}Urdu & \cellcolor{lblue}18.9  & \cellcolor{lblue}\multirow{-2}{*}{\textbf{17.7}} \\
        \hline
    \end{tabular}
    }
    \caption{Cross-modal verification results of the proposed method along with state-of-the-art methods on MAV-Celeb dataset. Bold: best, underlined: second best. }
    \label{tab:results_mavceleb}
\end{table}

    \noindent \textbf{Evaluation.}
    Following prior works~\cite{nagrani2018learnable,saeed2022fusion,saeed2023single}, we evaluate models' performance on cross-modal verification in terms of \textit{Equal Error Rate} (EER) and \textit{Area Under the Curve} (AUC), and on cross-modal matching in terms of \textit{Matching Accuracy}, with varying gallery size $n_c\in \{2, 4, 6, 8, 10\}$. 
    For the MAV-Celeb dataset, we report both the results separated by language (English or Urdu) and overall results. To analyze the impact of the inductive bias on the performance of our proposed model, we carry out an ablation study on both tasks of cross-modal verification and matching. In particular, we compare the performance of our proposed method with models relying only on cross-entropy loss $\mathcal{L}_\text{CE}$ during training ($\alpha=0$). For the model denoted with CE we do not include the fixed matrix imposing maximum class separation as inductive bias, for the one denoted with MSM we include the inductive-bias matrix~\cite{kasarla2022maximum}. Finally, the model denoted with FOP relies both on $\mathcal{L}_\text{CE}$ and $\mathcal{L}_\text{OC}$, but without the inductive-bias matrix. Notice that this model is the same as the one used by Saeed et al.~\cite{saeed2022fusion}, differing only in terms of the face encoder (FaceNet instead of VGGFace~\cite{parkhi2015deep}) and of the voice encoder (Ecapa-tdnn instead of Utterance Level Aggregation~\cite{xie2019utterance}).
    \vspace{-2mm}

\section{Results}
    \label{sec:results}
       \vspace{-0.7em}


\begin{figure}
    \centering
    \subfigure[]{\includegraphics[width=0.46\textwidth]{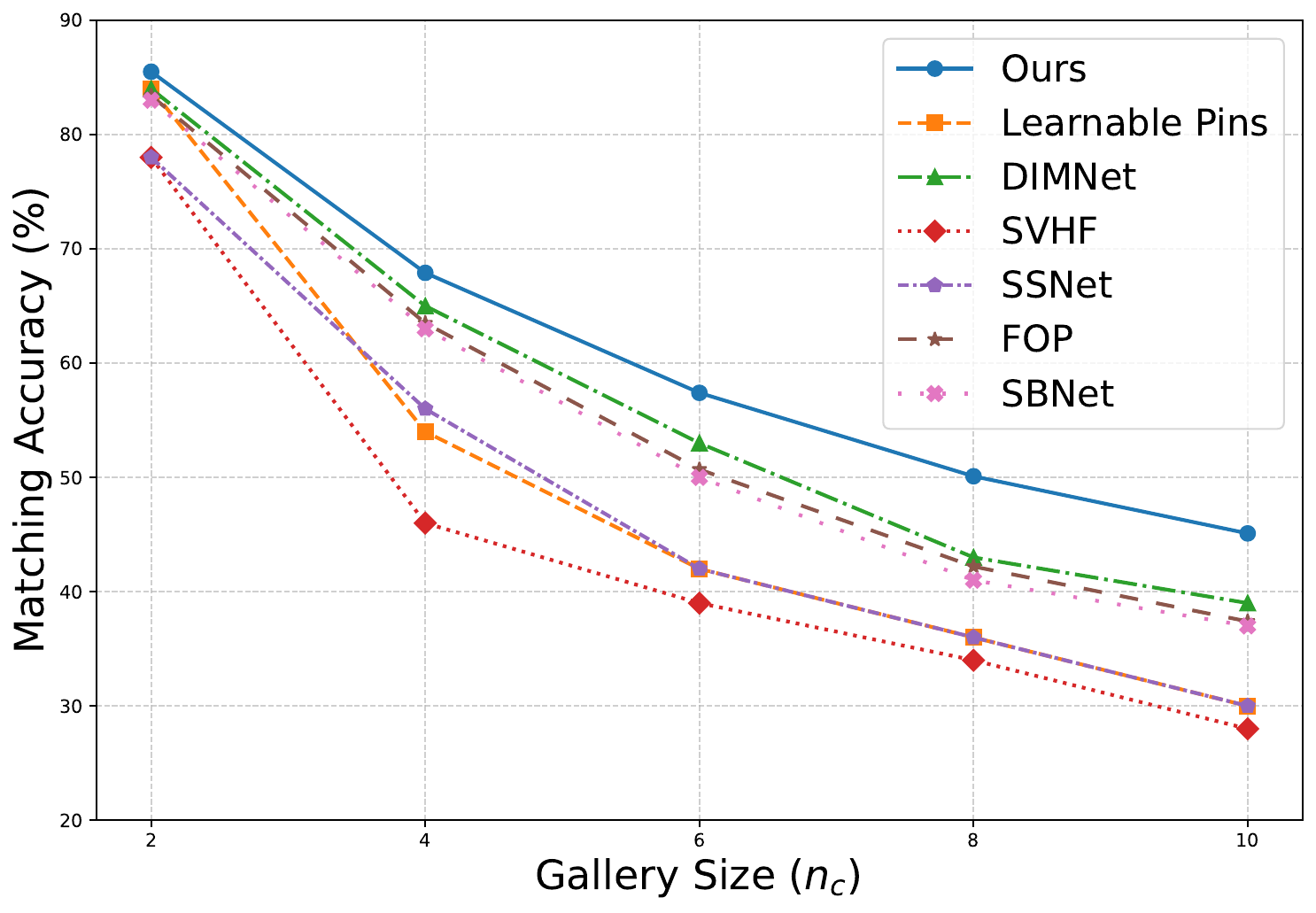}} 
    \label{subfig:matching_task}
    \subfigure[]{\includegraphics[width=0.46\textwidth]{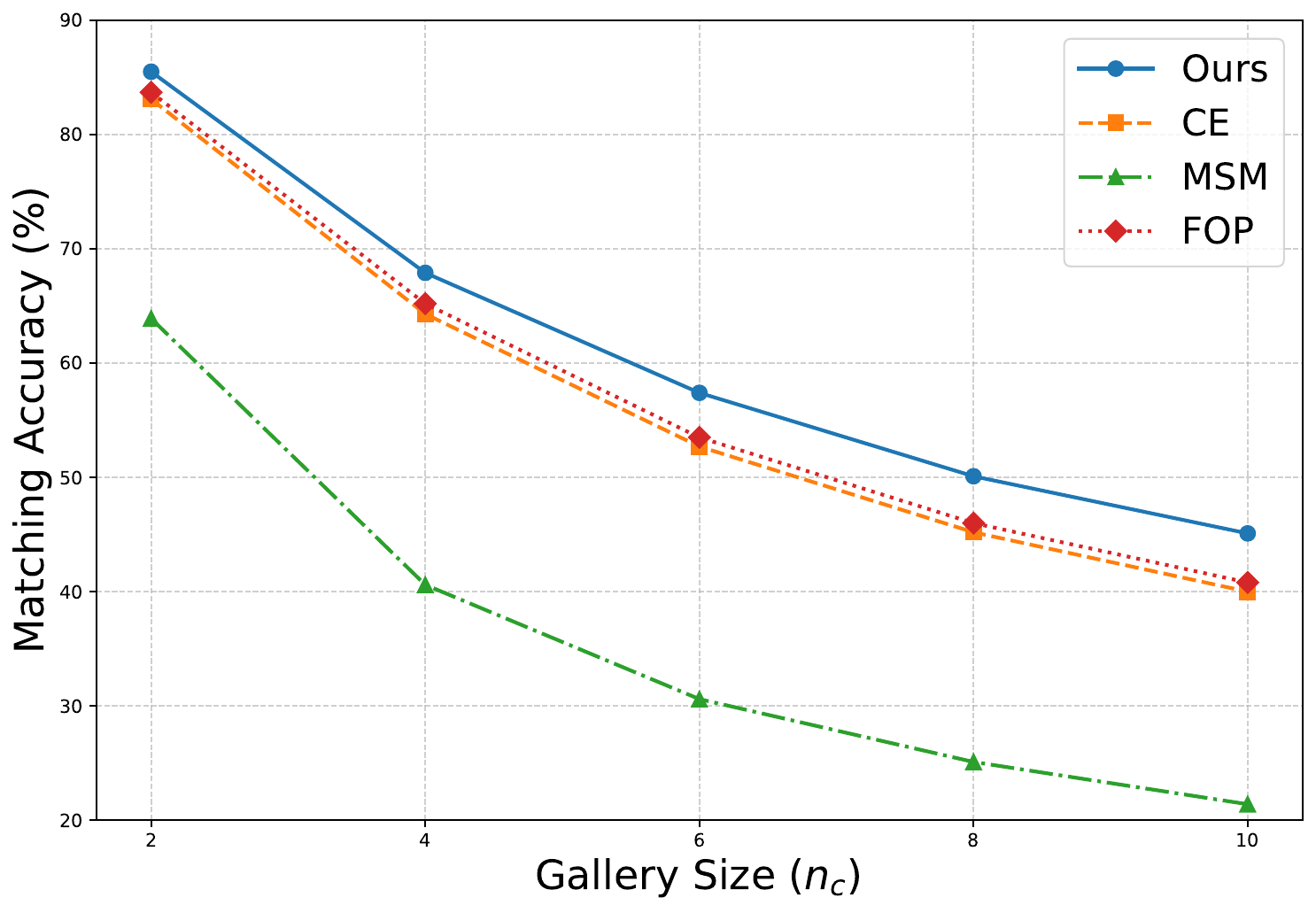}} 
    \label{subfigure:matching_ablation}
    \vspace{-6mm}
    \caption{Performance of the models for cross-modal matching on VoxCeleb 
    for varying gallery size $n_c\in\{2,4,6,8,10\}$. Left (a): comparison with SOTA methods. Right (b): ablation study on components of our method.}
    \label{fig:ablation_matching}
\end{figure}

    \noindent \textbf{Cross-Modal Verification.} 
    Table~\ref{tab:sota} shows the performance of the models on the task of cross-modal verification on the seen-heard and unseen-unheard configurations of VoxCeleb. The proposed method demonstrates superior performance and achieves SOTA results on both seen-heard and unseen-unheard scenarios and in terms of both EER and AUC.
    More specifically, on the seen-heard configuration our method reaches an EER of $13.9\%$, improving with respect to the previous SOTA result of the Single Stream Network~\cite{nawaz2019deep}  ($17.2\%$). On the unseen-unheard configuration, our model reaches a higher EER of $22.9\%$, which is to be expected, due to the harder configuration compared to seen-heard. However, even in this configuration our model reaches SOTA performance, outperforming the previous SOTA model FOP~\cite{saeed2022fusion}, that reaches an EER of $24.9\%$. The comparison between our model and FOP is particularly insightful, since the only difference in the architecture of these two models is the introduction of the inductive-bias matrix. The fact that our proposed model outperforms FOP in all metrics and on both configurations highlights the effectiveness of including maximum class separation as inductive bias in the architecture.    
    Table~\ref{tab:results_mavceleb} shows the EER on MAV-Celeb, and both for separate languages (Eng. or Urdu) and overall. Our model reaches the same performance of the SOTA model Audio-visual~\cite{chen2024contrastive} in terms of overall EER ($17.7\%$). Our model achieves SOTA on  English train ($16.5$), while performs lower than Audio-visual on Urdu language. 


\noindent \textbf{Cross-Modal Matching.} Figure~\ref{subfig:matching_task} a shows the performance of the models on cross-modal matching for VoxCeleb in terms of matching accuracy and for varying gallery size $n_c\in\{2,4,6,8,10\}$. 
All models show a decreasing accuracy as $n_c$ increases; this is 
expected because an increasing $n_c$ indicates a larger number of candidates
, for which models become less confident. Our proposed method (solid, blue with circles as markers) achieves a higher accuracy than all other SOTA methods and for all gallery sizes. The difference is more remarkable for higher gallery sizes, suggesting that our model has a higher matching confidence. 


    \noindent \textbf{Ablation Study}
    Table~\ref{tab:ablation_verification} shows the results of our ablation study on cross-modal verification task, comparing the performance of models for which different components are present, as described in Section~\ref{sec:method}. The same comparison is carried out on cross-modal matching task in Figure~\ref{subfigure:matching_ablation} b.
    We first observe that our model is most effective when all components ($\mathcal{L}_\text{CE}, \mathcal{L}_\text{OC}$ and inductive-bias matrix) are leveraged simultaneously. The second-best configuration is the one in which both loss functions are used during training, without the inductive-bias matrix. Interestingly, if trained only with $\mathcal{L}_\text{CE}$, the inductive-bias matrix degrades the performance (CE vs. MSM). 
    We attribute this to the fact that cross-modal verification and matching require embeddings to be maximally separated and compact, and hypothesize that for this reason MSM alone is not suitable for these tasks. 
    \vspace{-2mm}
    
    \begin{table}
    \centering
    \setlength{\tabcolsep}{16pt}
                    \scalebox{0.75}{
    \begin{tabular}{l|cc|cc}
    \rowcolor{Gray}
    \hline
    Method     & EER $\downarrow$ & AUC $\uparrow$ & EER $\downarrow$ & AUC $\uparrow$ \\
    \cline{2-5}
    \rowcolor{Gray}
     &  \multicolumn{2}{c|}{Seen-Heard}  & \multicolumn{2}{c}{Unseen-Unheard}\\
    \hline
    CE&  17.2 & 90.6 & 24.6 & 83.2 \\
    MSM              &  21.4 & 87.4  & 37.8 & 65.8   \\
    FOP$^*$ & \underline{16.7} & \underline{91.0} & \underline{23.8} & \underline{84.2}  \\
    \rowcolor{lblue}
    \hline
    Ours &  \textbf{13.9} & \textbf{93.7} & \textbf{22.9} & \textbf{85.0}  \\
    
    \hline
    \end{tabular}
    }
    \vspace{-4mm}
    \caption{Ablation study on VoxCeleb. FOP: no inductive-bias matrix. CE: no inductive-bias matrix and no $\mathcal{L}_\text{OC}$. MSM: no $\mathcal{L}_\text{OC}$. Bold: best, underlined: second best. Asterisk: different face and voice encoders than in Saeed et al.~\cite{saeed2022fusion}.}
    \label{tab:ablation_verification}
    \end{table}

    
\section{Conclusion}
    \vspace{-0.7em}
    In this work we proposed a method for face-voice association based on imposing an inductive bias to enforce maximum separation among speakers. 
    Our experiments showed that this method 
    establishes new SOTA performance on face-voice association
    . Being the first in applying maximum class separation as inductive bias in multimodal learning and beyond classification tasks, our work paves the way to establishing a new paradigm. There are, however, several limitations to the current work. We did not investigate how the performance improvement varies for a varying number of speakers. 
    We did not investigate the effectiveness of this approach on other  tasks. We applied maximum class separation as inductive bias to the previous SOTA model for face-voice association, 
    and did not investigate the impact of the inductive bias on other models.
    We leave these extensions for future research. 
    \vspace{-2mm}

\section{Acknowledgments}
   \vspace{-0.7em}
This research was funded in whole or in part by the Austrian Science Fund (FWF): Cluster of Excellence \href{https://www.bilateral-ai.net/home}{\textcolor{blue}{\textit{Bilateral Artificial Intelligence}}} (\url{https://doi.org/10.55776/COE12}) and the doc.funds.connect project \href{https://dfc.hcai.at/}{\textcolor{blue}{\textit{Human-Centered Artificial Intelligence}}} (\url{https://doi.org/10.55776/DFH23}). For open access purposes, the authors have applied a CC BY public copyright license to any author-accepted manuscript version arising from this submission.


\bibliographystyle{IEEEbib}
\small
\bibliography{strings,refs}

\end{document}